\documentclass[10pt,twocolumn,letterpaper]{article}

\usepackage{cvpr}
\usepackage{times}
\usepackage{epsfig}
\usepackage{graphicx}
\usepackage{amsmath}
\usepackage{amssymb}
\usepackage[hyphenbreaks]{breakurl}

\usepackage{multirow}
\usepackage{booktabs}
\usepackage{pifont}
\usepackage{xcolor}
\definecolor{red}{HTML}{FF3333}
\definecolor{blue}{HTML}{3399FF}
 \newcommand\hiddenref[1]{\sbox0{\ref{#1}}}

\usepackage[breaklinks=true,bookmarks=false]{hyperref}

\cvprfinalcopy 

\def\cvprPaperID{****} 

\begin{document}

\title{Multi-modal Dense Video Captioning}

\author{Vladimir Iashin \\
Tampere University \\
{\tt\small vladimir.iashin@tuni.fi}
\and
Esa Rahtu\\
Tampere University\\
{\tt\small esa.rahtu@tuni.fi}
}

\maketitle

\begin{abstract}
      Dense video captioning is a task of localizing interesting events from an untrimmed video and producing textual description (captions) for each localized event. Most of the previous works in dense video captioning are solely based on visual information and completely ignore the audio track. However, audio, and speech, in particular, are vital cues for a human observer in understanding an environment. In this paper, we present a new dense video captioning approach that is able to utilize any number of modalities for event description. Specifically, we show how audio and speech modalities may improve a dense video captioning model. We apply automatic speech recognition (ASR) system to obtain a temporally aligned textual description of the speech (similar to subtitles) and treat it as a separate input alongside video frames and the corresponding audio track. We formulate the captioning task as a machine translation problem and utilize recently proposed Transformer architecture to convert multi-modal input data into textual descriptions. We demonstrate the performance of our model on ActivityNet Captions dataset. The ablation studies indicate a considerable contribution from audio and speech components suggesting that these modalities contain substantial complementary information to video frames. Furthermore, we provide an in-depth analysis of the ActivityNet Caption results by leveraging the category tags obtained from original YouTube videos. Code is publicly available: \href{https://github.com/v-iashin/MDVC}{\texttt{\color{blue} github.com/v-iashin/MDVC}}.
\end{abstract}

\section{Introduction}

\begin{figure}[t]
\begin{center}
   \includegraphics[width=1\linewidth]{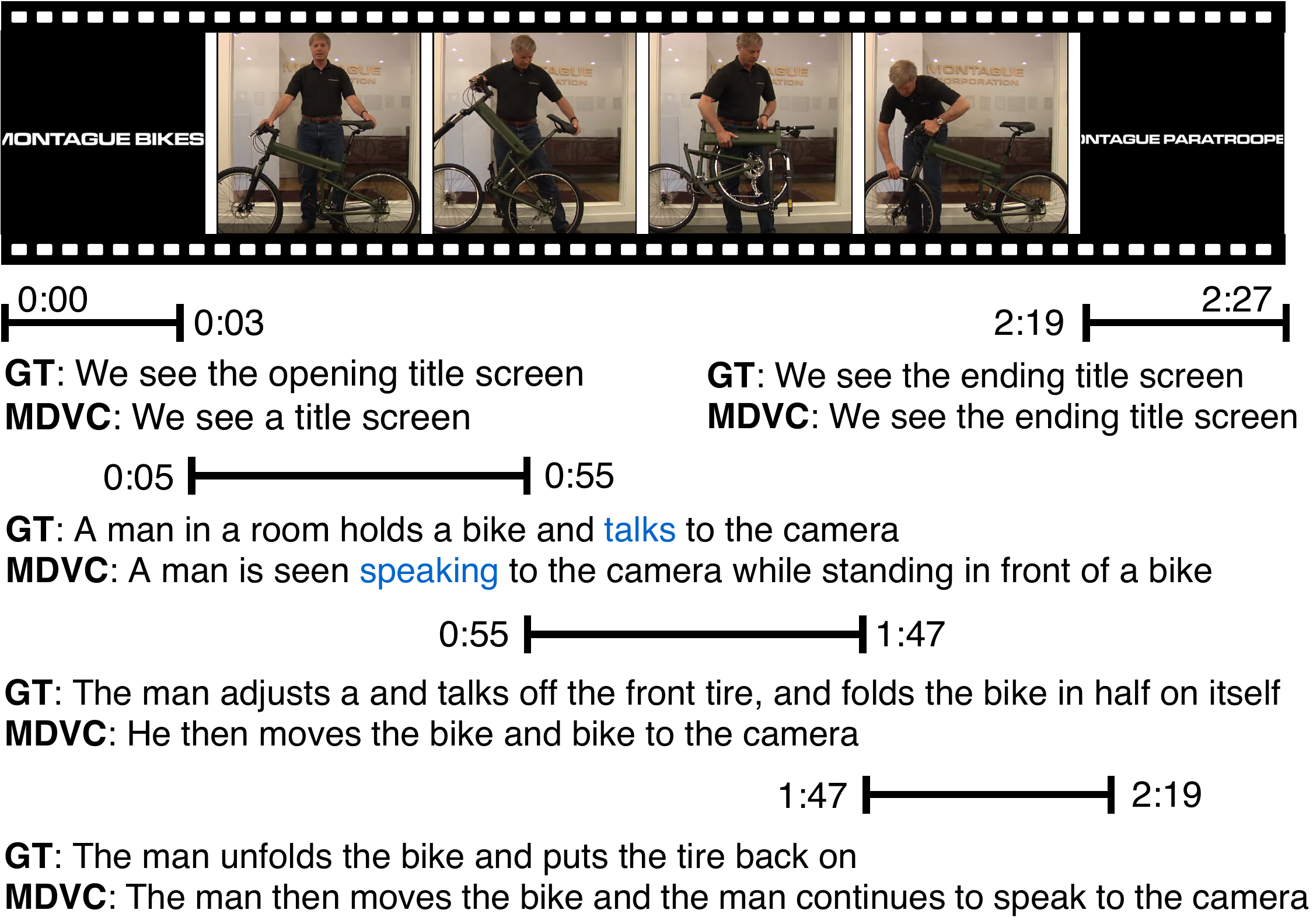}
\end{center}
   \caption{Example video with ground truth captions and predictions of Multi-modal Dense Video Captioning module (MDVC). It may account for any number of modalities, \ie audio or speech.\label{fig:intro}}
\end{figure}

The substantial amount of freely available video material has brought up the need for automatic methods to summarize and compactly represent the essential content. One approach would be to produce a short video skim containing the most important video segments as proposed in the \textit{video summarization} task \cite{Lee2012}. Alternatively, the video content could be described using natural language sentences. Such an approach can lead to a very compact and intuitive representation and is typically referred to as video captioning in the literature \cite{Yu2016}. However, producing a single description for an entire video might be impractical for long unconstrained footage. Instead, \textit{dense video captioning} \cite{Krishna2017} aims, first, at temporally localizing events and, then, at producing natural language description for each of them. Fig.~\ref{fig:intro} illustrates dense video captions for an example video sequence.

Most recent works in dense video captioning formulate the captioning problem as a machine translation task, where the input is a set of features extracted from the video stream and the output is a natural language sentence. Thus, the captioning methods can be leveraged by recent developments in machine translation field, such as \textit{Transformer} model \cite{Vaswani2017}. The main idea in the transformer is to utilise \textit{self-attention} mechanism to model long-term dependencies in a sequence. We follow the recent work \cite{Zhou2018} and adopt the transformer architecture in our dense video captioning model. 

The vast majority of previous works are generating captions purely based on visual information \cite{Zhou2018,Wang2018n,Li2018,Ma2018,Xiong2018,Mun2019,Xu2019}. However, almost all videos include an audio track, which could provide vital cues for video understanding. In particular, what is being said by people in the video, might make a crucial difference to the content description. For instance, in a scene when someone knocks the door from an opposite side, we only see the door but the audio helps us to understand that somebody is behind it and wants to enter. Therefore, it is impossible for a model to make a useful caption for it. Also, other types of videos as instruction videos, sport videos, or video lectures could be challenging for a captioning model. 

In contrast, we build our model to utilize video frames, raw audio signal, and the speech content in the caption generation process. To this end, we deploy \textit{automatic speech recognition} (ASR) system \cite{YouTube} to extract time-aligned captions of \textit{what is being said} (similar to subtitles) and employ it alongside with video and audio representations in the transformer model. 

The proposed model is assessed using the challenging ActivityNet Captions \cite{Krishna2017} benchmark dataset, where we obtain competitive results to the current state-of-the-art. The subsequent ablation studies indicate a substantial contribution from audio and speech signals. Moreover, we retrieve and perform breakdown analysis by utilizing previously unused video category tags provided with the original YouTube videos \cite{YouTubeCats}. The program code of our model and the evaluation approach will be made publicly available. 

\section{Related Work}

\subsection{Video Captioning}

Early works in video captioning applied \textit{rule-based models} \cite{Kojima2002,MunWaiLee2008,Das2013}, where the idea was to identify a set of video objects and use them to fill predefined templates to generate a sentence. Later, the need for sentence templates was omitted by casting the captioning problem as a machine translation task \cite{Rohrbach2013}. Following the success of neural models in translation systems \cite{Sutskever2014}, similar methods became widely popular in video captioning \cite{Yao2015,Venugopalan2015,Venugopalan2015b,Yu2016,Baraldi2017,Shen2017,Hori2017,Donahue2017,Wu2018}. The rationale behind this approach is to train two \textit{Recurrent Neural Networks} (RNNs) in an \textit{encoder-decoder} fashion. Specifically, an encoder inputs a set of video features, accumulates its \textit{hidden state}, which is passed to a decoder for producing a caption. 

To further improve the performance of the captioning model, several methods have been proposed, including shared memory between visual and textual domains \cite{Wang2018a,Pei2019}, spatial and temporal attention \cite{Yan2019}, reinforcement learning \cite{Wang2018}, semantic tags \cite{Gan2017,Pan2017}, other modalities \cite{Xu2017,Hori2018,Wang2018b,Hao2017}, and by producing a paragraph instead of one sentence \cite{Rohrbach2014,Yu2016}.

\subsection{Dense Video Captioning}

Inspired by the idea of the \textit{dense image captioning} task \cite{Johnson2016}, Krishna \etal \cite{Krishna2017} introduced a problem of dense video captioning and released a new dataset called ActivityNet Captions which leveraged the research in the field \cite{Zhou2018,Wang2018n,Li2018,Ma2018,Xiong2018,Mun2019,Rahman2019,Xu2019}. In particular, \cite{Wang2018n} adopted the idea of the context-awareness \cite{Krishna2017} and generalized the temporal event proposal module to utilize both past and future contexts as well as an \textit{attentive fusion} to differentiate captions from highly overlapping events. Meanwhile, the concept of \textit{Single Shot Detector} (SSD) \cite{Liu2016} was also used to generate event proposals and \textit{reward maximization} for better captioning in \cite{Li2018}. 

In order to mitigate the intrinsic difficulties of RNNs to model long-term dependencies in a sequence, Zhou \etal \cite{Zhou2018} tailored the recent idea of Transformer \cite{Vaswani2017} for dense video captioning. In \cite{Ma2018} the authors noticed that the captioning may benefit from interactions between objects in a video and developed recurrent higher-order interaction module to model these interactions. Xiong \etal \cite{Xiong2018} noticed that many previous models produced redundant captions, and proposed to generate captions in a progressive manner, conditioned on the previous caption while applying paragraph- and sentence-level rewards. Similarly, a ``bird-view'' correction and two-level reward maximization for a more coherent story-telling have been employed in \cite{Mun2019}.

Since the human annotation of a video with temporal boundaries and captions for each of them can be laborious, several attempts have been made to address this issue \cite{Xuguang2018,Miech2019}. Specifically, \cite{Xuguang2018} employed the idea of \textit{cycle-consistency} to translate a set of captions to a set of temporal events without any paired annotation, while \cite{Miech2019} automatically-collected dataset of an unparalleled-scale exploiting the structure of instructional videos. 

The most similar work to our captioning model is \cite{Zhou2018} that also utilizes a version of the Transformer \cite{Vaswani2017} architecture. However, their model is designed solely for visual features. Instead, we believe that dense video captioning may benefit from information from other modalities. 

\subsection{Multi-modal Dense Video Captioning}

A few attempts has been made to include additional cues like audio and speech \cite{Rahman2019,ACase_Hessel_2019,DenseProcCap_Shi_2019} for dense video captioning task. Rahman \etal~\cite{Rahman2019} utilized the idea of \textit{cycle-consistency} \cite{Xuguang2018} to build a model with visual and audio inputs. However, due to weak supervision, the system did not reach high performance. Hessel \etal~\cite{ACase_Hessel_2019} and Shi \etal~\cite{DenseProcCap_Shi_2019} employ a transformer architecture \cite{Vaswani2017} to encode both video frames and speech segments to generate captions for instructional (cooking) videos. Yet, the high results on a dataset which is restricted to instructional video appear to be not evidential as the speech and the captions are already very close to each other in such videos \cite{Miech2019}. 

\begin{figure*}[t!]
    \begin{center}
    \includegraphics[width=1\linewidth]{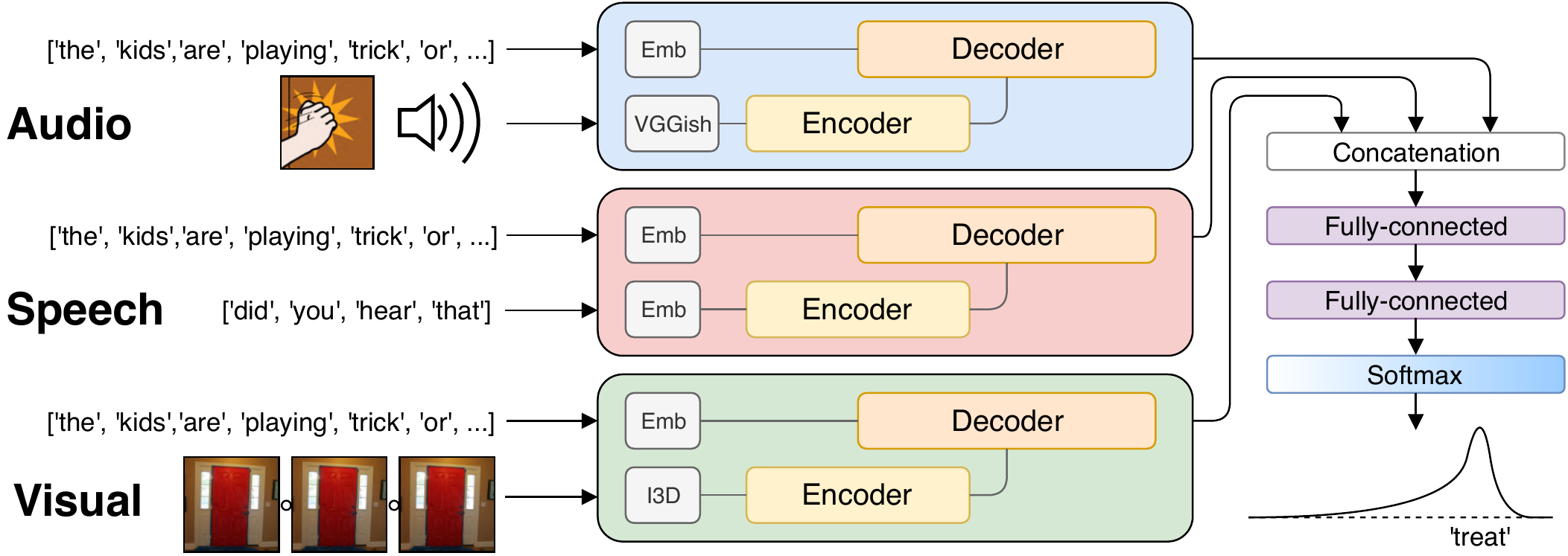}
    \end{center}
    \caption{The proposed Multi-modal Dense Video Captioning (MDVC) framework. Given an input consisting of several modalities, namely, audio, speech, and visual, internal representations are produced by a corresponding feature transformer (middle). Then, the features are fused in the multi-modal generator (right) that outputs the distribution over the vocabulary.\label{fig:mdvc}}
\end{figure*}

\begin{figure*}[t!]
    \begin{center}
    \includegraphics[width=1\linewidth]{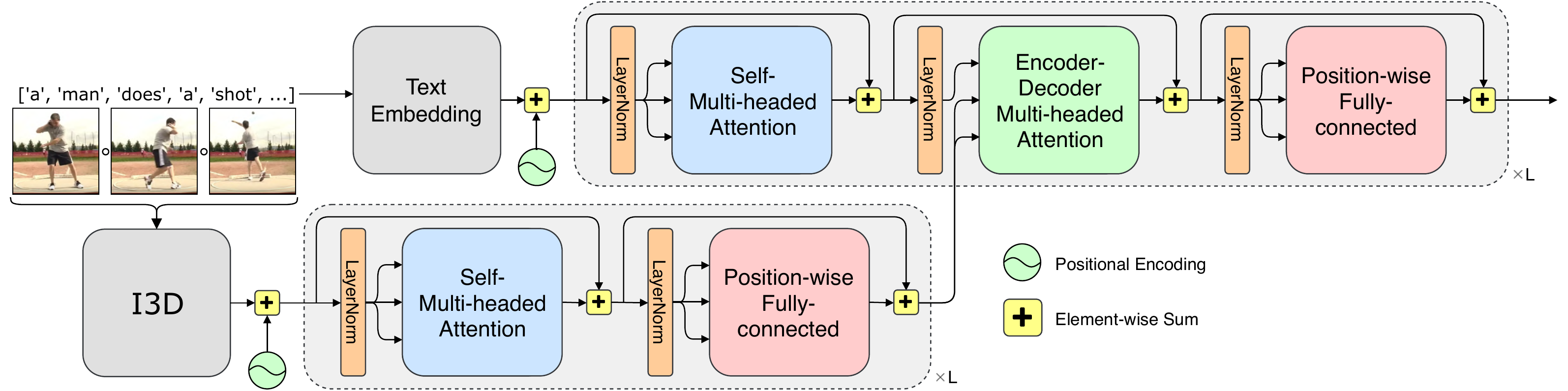}
    \end{center}
    \caption{The proposed feature transformation architecture that consists of an encoder (bottom part) and a decoder (top part). The encoder inputs pre-processed and position-encoded features from I3D (in case of the visual modality), and outputs an internal representation. The decoder, in turn, is conditioned on both position-encoded caption that is generated so far and the output of the encoder. Finally, the decoder outputs its internal representation.\label{fig:transformer}}
\end{figure*}

In contrast to the mentioned multi-modal dense video captioning methods: (1) we present the importance of the speech and audio modalities on a domain-free dataset, (2) propose a multi-modal dense video captioning module (MDVC) which can be scaled to any number of modalities. \hiddenref{fig:mdvc} \hiddenref{fig:transformer}

\section{Proposed Framework}
In this section, we briefly outline the workflow of our method referred to as Multi-modal Dense Video Captioning (MDVC) which is shown in Fig.~\ref{fig:mdvc}. The goal of our method is to temporally localize events on a video and to produce a textual description for each of them. To this end, we apply a two-stage approach. 

Firstly, we obtain the temporal event locations. For this task, we employ the \textit{Bidirectional Single-Stream Temporal} action proposals network (Bi-SST) proposed in \cite{Wang2018n}. Bi-SST applies 3D Convolution network (C3D) \cite{Tran2015} to video frames and extracts features that are passed to subsequent \textit{bi-directional LSTM} \cite{Hochreiter1997} network. The LSTM accumulates visual cues over time and predicts confidence scores for each location to be start/end point of an event. Finally, a set of event proposals (start/end times) is obtained and passed to the second stage for caption generation. 

Secondly, we generate the captions given a proposal. To produce inputs from audio, visual, and speech modalities, we use \textit{Inflated 3D convolutions} (I3D) \cite{Carreira2017} for visual and \textit{VGGish network} \cite{Hershey2017} for audio modalities. For speech representation as a text, we employ an external ASR system \cite{YouTube}. To represent the text into a numerical form, we use a similar text embedding which is used for caption encoding. The features are, then, fed to individual transformer models along with the words of a caption from the previous time steps. The output of the transformer is passed into a \textit{generator} which fuses the outputs from all modalities and estimates a probability distribution over the word vocabulary. After sampling the next word, the process is repeated until a special end \textit{token} is obtained. Fig.~\ref{fig:intro} illustrates an example modality and the corresponding event captions. 

\subsection{Temporal Event Localization Module}
An event localization module is dedicated to generating a set of temporal regions which might contain an event. To achieve this, we employ pre-trained Bidirectional Single-Stream Temporal action proposals network (Bi-SST) proposed in \cite{Wang2018n} as it has is been shown to reach good performance in the proposal generation task.

Bi-SST inputs a sequence of $T$ RGB frames from a video $V = (x_1, x_2, \dots, x_F)$ and extracts a set of $4096$-d features $V' = (f_1, f_2, \dots, f_T)$ by applying a 3D Convolution network (C3D) on non-overlapping segments of size 16 with a stride of 64 frames. To reduce the feature dimension, only $500$ principal components were selected using PCA.

To account for the video context, events are proposed during forward and backward passes on a video sequence $V'$, and, then, the resulting scores are fused together to obtain the final proposal set. Specifically, during the \textit{forward pass}, LSTM is used to accumulate the visual clues from the ``past'' context at each position $t$ which is treated as an \textit{ending} point and produce confidence scores for each proposal.

Afterwards, a similar procedure is performed during the \textit{backward pass} where the features $V'$ are used in a reversed order. This empowers the model to have a sense of the ``future'' context in a video. In contrast to the forward pass, each position is treated as a \textit{starting} point of the proposal. Finally, the confidence scores from both passes are fused by multiplication of corresponding scores for each proposal at each time step, and, then, filtered according to a predefined threshold.

Finally, we obtain a set of $N_V$ event proposals for caption generation $P_V=\{p_j = (\text{start}_j, \text{end}_j, \text{score}_j)\}_{j=1}^{N_V}$.

\subsection{Captioning Module}
In this section we explain the captioning based for an example modality, namely, visual. Given a video $V$ and a set of proposals $P_V$ from the event localization module, the task of the captioning module is to provide a caption for each proposal in $P_V$. In order to extract features from a video $V$, we employ I3D network \cite{Carreira2017} pre-trained on the Kinetics dataset which produces $1024$-d features. The gap between the extracted features and the generated captions is filled with Transformer \cite{Vaswani2017} architecture which was proven to effectively encode and decode the information in a sequence-to-sequence setting. 

\subsubsection{Feature Transformer}
As shown in Fig.~\ref{fig:transformer}, Feature Transformer architecture mainly consists of three blocks: an \textit{encoder}, \textit{decoder}, and \textit{generator}. The encoder inputs a set of extracted features $ \mathbf{v}^j = (v_1, v_2, \dots, v_{T_j}) $ temporally corresponding to a proposal $p_j$ from $P_V$ and maps it to a sequence of internal representations $ \mathbf{z}^j = (z_1, z_2, \dots, z_{T_j}) $. The decoder is conditioned on the output of the encoder $\mathbf{z}^j$ and the \textit{embedding} $ \mathbf{e}^j_{\leqslant t} = (e_1, e_2, \dots, e_t)$ of the words in a caption $ \mathbf{w}^j_{\leqslant t} = (w_1, w_2, \dots, w_t) $. It produces the representation $ \mathbf{g}^j_{\leqslant t} = (g_1, g_2, \dots, g_t) $ which, in turn, is used by the generator to model a distribution over a vocabulary for the next word $ p(w_{t+1}|\mathbf{g}^j_{\leqslant t}) $. The next word is selected greedily by obtaining the word with the highest probability until a special \textit{ending token} is sampled. The captioning is initialized with a \textit{starting token}. Both are added to the vocabulary.

Before providing an overview of the encoder, decoder, and generator, we presenting the notion of \textit{multi-headed attention} that acts as an essential part of the decoder and encoder blocks. The concept of the multi-head attention, in turn, heavily relies on \textit{dot-product attention} which we describe next.

\paragraph{Dot-product Attention}
The idea of the multi-headed attention rests on the \textit{scaled dot-product attention} which calculates the weighted sum of \textit{values}. The weights are obtained by applying the softmax function on the dot-product of each pair of rows of \textit{queries} and \textit{keys} scaled by $\frac{1}{\sqrt{D_k}}$. The scaling is done to prevent the softmax function from being in the small gradient regions \cite{Vaswani2017}. Formally the scaled dot-product attention can be represented as follows
\begin{align}\label{eq:attention}
    \text{Attention}(Q, K, V) = \text{softmax}\Bigg(\frac{QK^T}{\sqrt{D_k}}\Bigg)V,
\end{align}
where $Q, K, V $ are queries, keys, and values, respectively.

\paragraph{Multi-headed Attention} 
The multi-headed attention block is used once in each encoder layer and twice in each decoder layer. The block consists of $H$ \textit{heads} that allows to cooperatively account for information from several representations sub-spaces at every position while preserving the same computation complexity \cite{Vaswani2017}. In a transformer with dimension $D_T$, each head is defined in the following way
\begin{align}
    \text{head}_h(q, k, v) = \text{Attention}(qW^{q}_h, kW^{k}_h, vW^{v}_h),
\end{align}
where $q, k, v$ are matrices which have $D_T$ columns and the number of rows depending on the position of the multi-headed block, yet with the same number of rows for $k$ and $v$ to make the calculation in \eqref{eq:attention} to be feasible. The $W^{q}_h, W^{k}_h, W^{v}_h \in \mathbb{R}^{D_T \times D_k}$ are trainable projection matrices that map $q, k , v$ from $D_T$ into $D_k= \frac{D_T}{H}$, asserting $D_T$ is a multiple of $H$.  The multi-head attention, in turn, is the concatenation of all attention heads mapped back into $D_T$ by trainable parameter matrix $W^o \in \mathbb{R}^{D_k \cdot H \times D_T}$:
\begin{align}
    \text{MultiHead}(q, k, v) = \begin{bmatrix}
                                    \text{head}_1(q, k, v) \\
                                    \dots \\
                                    \text{head}_H(q, k, v)
                                \end{bmatrix} W^o.
\end{align}

\paragraph{Encoder} 
The encoder consists of $ L $ layers. The first layer inputs a set of features $ \mathbf{v}^j $ and outputs an internal representation  $ \mathbf{z}_1^j \in \mathbb{R}^{T_j \times D_T} $ while each of the next layers treats the output of a previous layer as its input. Each encoder layer $l$ consist of two sub-layers: \textit{multi-headed attention} and \textit{position-wise fully connected network} which are explained later in this section. The input to both sub-layers are normalized using layer normalization \cite{Ba2016}, each sub-layer is surrounded by a residual connection \cite{He2016} (see Fig.~\ref{fig:transformer}). Formally, the $l$-th encoder layer has the following definition
\begin{align}
    \overline{\mathbf{z}}_l^j &= \text{LayerNorm}(\mathbf{z}_l^j) \\
    \mathbf{r}^j_l &= \mathbf{z}_l^j + \text{MultiHead}( \overline{\mathbf{z}}_l^j, \overline{\mathbf{z}}_l^j, \overline{\mathbf{z}}_l^j) \\
    \overline{\mathbf{r}}_l^j &= \text{LayerNorm}(\mathbf{r}^j_l) \\
    \mathbf{z}_{l+1}^j &= \mathbf{r}_l^j + \text{FCN}(\overline{\mathbf{r}}_l^j),
\end{align}
where $\text{FCN}$ is the position-wise fully connected network. Note, the multi-headed attention has identical queries, keys, and values ($ \overline{\mathbf{z}}_l^j $). Such multi-headed attention block is also referred to as \textit{self}-multi-headed attention. It enables an encoder layer $l$ to account for the information from all states from the previous layer $ \mathbf{z}_{l-1}^j$. This property contrasts with the idea of RNN which accumulates only the information from the past positions. 

\paragraph{Decoder} 
Similarly to the encoder, the decoder has $ L $ layers. At a position $t$, the decoder inputs a set of embedded words $\mathbf{e}^j_{\leqslant t}$ with the output of the encoder $\mathbf{z}^j$ and sends the output to the next layer which is conditioned on this output and, again, the encoder output $\mathbf{z}^j$. Eventually, the decoder producing its internal representation $\mathbf{g}_{\leqslant t}^j \in \mathbb{R}^{t \times D_T}$. The decoder block is similar to the encoder but has an additional sub-layer that applies multi-headed attention on the encoder output and the output of its previous sub-layer. The decoder employs the layer normalization and residual connections at all three sub-layers in the same fashion as the encoder. Specifically, the $l$-th decoder layer has the following form:
\begin{align}
    \overline{\mathbf{g}}_l^j &= \text{LayerNorm}(\mathbf{g}_{l,\leqslant t}^j) \\
    \mathbf{b}^j_{l} &= \mathbf{g}_{l,\leqslant t}^j + \text{MultiHead}( \overline{\mathbf{g}}_l^j, \overline{\mathbf{g}}_l^j, \overline{\mathbf{g}}_l^j) \label{eq:selfatt_decoder}\\
    \overline{\mathbf{b}}_l^j &= \text{LayerNorm}(\mathbf{b}^j_{l}) \\
    \mathbf{u}^j_{l} &= \mathbf{g}_{l,\leqslant t}^j + \text{MultiHead}( \overline{\mathbf{b}}_l^j, \mathbf{z}^j, \mathbf{z}^j) \\
    \overline{\mathbf{u}}_l^j &= \text{LayerNorm}(\mathbf{u}^j_{l}) \\
    \mathbf{g}_{l+1, \leqslant t}^j &= \mathbf{u}^j_{l} + \text{FCN}(\overline{\mathbf{u}}_l^j),
\end{align}
where $ \mathbf{z}^j $ is the encoder output. Note, similarly to the encoder, \eqref{eq:selfatt_decoder} is a self-multi-headed attention function while the second multi-headed attention block attends on both the encoder and decoder and is also referred to as \textit{encoder-decoder} attention. This block enables each layer of the decoder to attend all state of the encoder's output $ \mathbf{z}^j$.

\paragraph{Position-wise Fully-Connected Network} 
The fully connected network is used in each layer of the encoder and the decoder. It is a simple two-layer neural network that inputs $x$ with the output of the multi-head attention block, and, then, projects each row (or position) of the input $x$ from $D_T$ space onto $D_P$, $(D_P > D_T)$ and back, formally:
\begin{align}
    \text{FCN}(x) = \text{ReLU}(xW_1 + b_1)W_2 + b_2,
\end{align}
where $W_1 \in \mathbb{R}^{D_T \times D_P}$, $W_2 \in \mathbb{R}^{D_P \times D_T}$, and biases $b_1, b_2$ are trainable parameters, $\text{ReLU}$ is a rectified linear unit.

\paragraph{Generator} 
At the position $t$, the generator consumes the output of the decoder $\mathbf{g}^j_{\leqslant t}$ and produces a distribution over the vocabulary of words $p(w_{t+1}| \mathbf{g}^j_{\leqslant t})$. To obtain the distribution, the generator applies the softmax function of the output of a fully connected layer with a weight matrix $W_G \in \mathbb{R}^{D_T \times D_V}$ where $D_V$ is a vocabulary size. The word with the highest probability is selected as the next one.

\paragraph{Input Embedding and Positional Encoding} 
Since the representation of textual data is usually sparse due to a large vocabulary, the dimension of the input of a neural language model is reduced with an embedding into a dimension of a different size, namely $D_T$. Also, following \cite{Vaswani2017}, we multiply the embedding weights by $\sqrt{D_T}$.  The \textit{position encoding} is required to allow the transformer to have a sense of the order in an input sequence. We adopt the approach proposed for a transformer architecture, \ie we add the output of the combination of sine and cosine functions to the embedded input sequence \cite{Vaswani2017}. 

\subsubsection{Multi-modal Dense Video Captioning} 
In this section, we present the multi-modal dense video captioning module which, utilises visual, audio, and speech modalities. See Fig.~\ref{fig:transformer} for a schematic representation of the module.

For the sake of speech representation $\mathbf{s}^j = (s_1, s_2, \dots, s_{T_j^s})$, we use the text embedding of size $512$-d that is similar to the one which is employed in the embedding of a caption $\mathbf{w}^j_{\leqslant t}$. To account for the audio information, given a proposal $p_j$ we extract a set of features $\mathbf{a}_j = (a_1, a_2, \dots, a_{T_j^a})$ applying the $128$-d embedding layer of the pre-trained VGGish network \cite{Hershey2017} on an audio track. While the visual features $\mathbf{v}^j = (v_1, v_2, \dots v_{T_j^v}) $ are encoded with $1024$-d vectors by Inflated 3D (I3D) convolutional network \cite{Carreira2017}.

To fuse the features, we create an encoder and a decoder for each modality with dimensions corresponding to the size of the extracted features. The outputs from all decoders are fused inside of the generator, and the distribution of a next word $w_{t+1}$ is formed. 

In our experimentation, we found that a simple two-layer fully-connected network applied of a matrix of concatenated features performs the best with the \textit{ReLU} activation after the first layer and the softmax after the second one. Each layer of the network has a matrix of trainable weights: $W_{F_1} \in \mathbb{R}^{D_F \times D_V}$ and $W_{F_2} \in \mathbb{R}^{D_V \times D_V}$ with $D_F = 512 + 128 + 1024 $ and $D_V$ is a vocabulary size.

\subsection{Model Training}
As the training is conducted using mini-batches of size 28, the features in one modality must be of the same length so the features could be stacked into a tensor. In this regard, we \textit{pad} the features and the embedded captions to match the size of the longest sample. 

The model is trained by optimizing the Kullback--Leibler divergence loss which measures the ``distance'' between the ground truth and predicted distributions and averages the values for all words in a batch ignoring the \textit{masked} tokens. 

Since many words in the English language may have several synonyms or human annotation may contain mistakes, we undergo the model to be less certain about the predictions and apply Label Smoothing \cite{Szegedy2016} with the smoothing parameter $\gamma$ on the ground truth labels to mitigate this. In particular, the ground truth distribution over the vocabulary of size $D_V$, which is usually represented as one-hot encoding vector, the identity is replaced with probability $1-\gamma$ while the rest of the values are filled with $\frac{\gamma}{D_V-1}$.

During training, we exploit the \textit{teacher forcing} technique which uses the ground truth sequence up to position $t$ as the input to predict the next word instead of using the sequence of predictions. As we input the whole ground truth sequence at once and predicting the next words at each position, we need to prevent the transformer from peeping for the information from the next positions as it attends to all positions of the input. To mitigate this, we apply masking inside of the self-multi-headed attention block in the decoder for each position higher than $t-1$, following \cite{Vaswani2017}.

The details on the feature extraction and other implementation details are available in the supplementary materials.

\begin{table}[t]
\centering
\small{
\centering
\setlength\tabcolsep{0.41em} 
\begin{tabular}{llrrrrr}
\toprule
\multirow{2}{*}{Method}          & \multicolumn{3}{c}{GT Proposals}  & \multicolumn{3}{c}{Learned Proposals}    \\
                                    & B@3  & B@4   & M & B@3  & B@4   & M  \\ \midrule \midrule
\multicolumn{2}{l}{\textbf{\textit{Seen full dataset}}}    &  & &  &  & \\
Krishna \etal \cite{Krishna2017} & 4.09 & 1.60 & 8.88  & 1.90 & 0.71 & 5.69 \\
Wang \etal \cite{Wang2018n}      & --   & --   & 10.89 & 2.55 & \textbf{1.31} & 5.86 \\
Zhou \etal \cite{Zhou2018}       & 5.76 & 2.71 & 11.16 & 2.42 & 1.15 & 4.98 \\
Li \etal \cite{Li2018}           & 4.55 & 1.62 & 10.33 & 2.27 & 0.73 & 6.93 \\ \midrule \midrule
\multicolumn{2}{l}{\textbf{\textit{Seen part of the dataset}}}  &  & &  &  & \\
Rahman \etal \cite{Rahman2019}   & 3.04 & 1.46 & 7.23  & 1.85 & 0.90 & 4.93 \\
MDVC   & 4.12 & 1.81 & 10.09 & 2.31 & 0.92 & 6.80   \\
MDVC, no missings       & \textbf{5.83} & \textbf{2.86} & \textbf{11.72} & \textbf{2.60} & 1.07 & \textbf{7.31} \\ \bottomrule
\end{tabular}
}
\caption{
The results of the dense video captioning task on the ActivityNet Captions validation sets in terms of BLEU--3,4 (B@3, B@4) and METEOR (M). The related methods are compared with the proposed approach (MDVC) in two settings: on the full validation dataset and a part of it with the videos with all modalities present for a fair comparison (``no missings''). Methods are additionally split into the ones which ``saw'' all training videos and another ones which trained on partially available data. The results are presented for both ground truth (GT) and learned proposals.\label{tab:captioning}
}
\end{table}

\section{Experiments}

\subsection{Dataset}
We perform our experiments using ActivityNet Captions dataset \cite{Krishna2017} that is considered as the standard benchmark for dense video captioning task. The dataset contains approximately 20k videos from YouTube and split into 50/25/25\,\% parts for training, validation, and testing, respectively. Each video, on average, contains 3.65 temporally localized captions, around 13.65 words each, and two minutes long. In addition, each video in the validation set is annotated twice by different annotators. We report all results using the validation set (no ground truth is provided for the test set). 

The dataset itself is distributed as a collection of links to YouTube videos, some of which are no longer available. Authors provide pre-computed C3D features and frames at 5fps, but these are not suitable for our experiments. At the time of writing, we found 9,167 (out of 10,009) training and 4,483 (out of 4,917) validation videos which is, roughly, 91\,\% of the dataset. Out of these 2,798 training and 1,374 validation videos (approx. 28\,\%) contain at least one speech segment. The speech content was obtained from the \textit{closed captions} (CC) provided by the YouTube ASR system which can be though as subtitles.

\subsection{Metrics}\label{sec:metrics}

We are evaluating the performance of our model using BLEU@N \cite{Papineni2002} and METEOR \cite{Denkowski2014}. We regard the METEOR as our primary metric as it has been shown to be highly correlated with human judgement in a situation with a limited number of references (only one, in our case). 

We employ the official evaluation script provided in \cite{KrishnaGithub}. Thus, the metrics are calculated if a proposed event and a ground truth location of a caption overlaps more than a specified \textit{temporal Intersection over Union} (tIoU) and zero otherwise. All metric values are averaged for every video, and, then, for every threshold tIoU in $[0.3, 0.5, 0.7, 0.9]$. On the validation, we average the resulting scores for both validation sets. For the learned proposal setting, we report our results on at most 100 proposals per video.

Notably, up to early 2017, the evaluation code had an issue which previously overestimated the performance of the algorithms in the learned proposal setting \cite{Mun2019}. Therefore, we report the results using the new evaluation code. 

\subsection{Comparison with Baseline Methods}
We compare our method with five related approaches, namely Krishna \etal \cite{Krishna2017}, Wang \etal \cite{Wang2018n}, Zhou \etal \cite{Zhou2018}, Li \etal \cite{Li2018}, and Rahman \etal \cite{Rahman2019}. We take the performance values from the original papers, except for \cite{Li2018}, and \cite{Zhou2018}, which are taken from \cite{Mun2019} due to the evaluation issue (see Sec.~\ref{sec:metrics}). 

The lack of access to the full ActivityNet Captions dataset makes strictly fair comparison difficult as we have less training and validation videos. Nevertheless, we present our results in two set-ups: 1) full validation set with random input features for missing entries, and 2) videos with all three modalities present (video, audio, and speech). The first one is chosen to indicate the lower bound of our performance with the full dataset. Whereas, the second one (referred to as ``\textit{no missings}'') concentrates on the multi-modal setup, which is the main contribution of our work. 

The obtained results are presented in Tab.~\ref{tab:captioning}. Our method (MDVC) achieves comparable or better performance, even though we have access to smaller training set and $9$\,\% of the validation videos are missing (replaced with random input features). Furthermore, if all three modalities are present, our method outperforms all baseline approaches in the case of both GT and learned proposals. Notably, we outperform \cite{Zhou2018} which is also based on the transformer architecture and account for the optical flow. This shows the superior performance of our captioning module which, yet, trained on the smaller amount of data. 

\begin{figure*}[t]
    \begin{center}
    \includegraphics[width=1\linewidth]{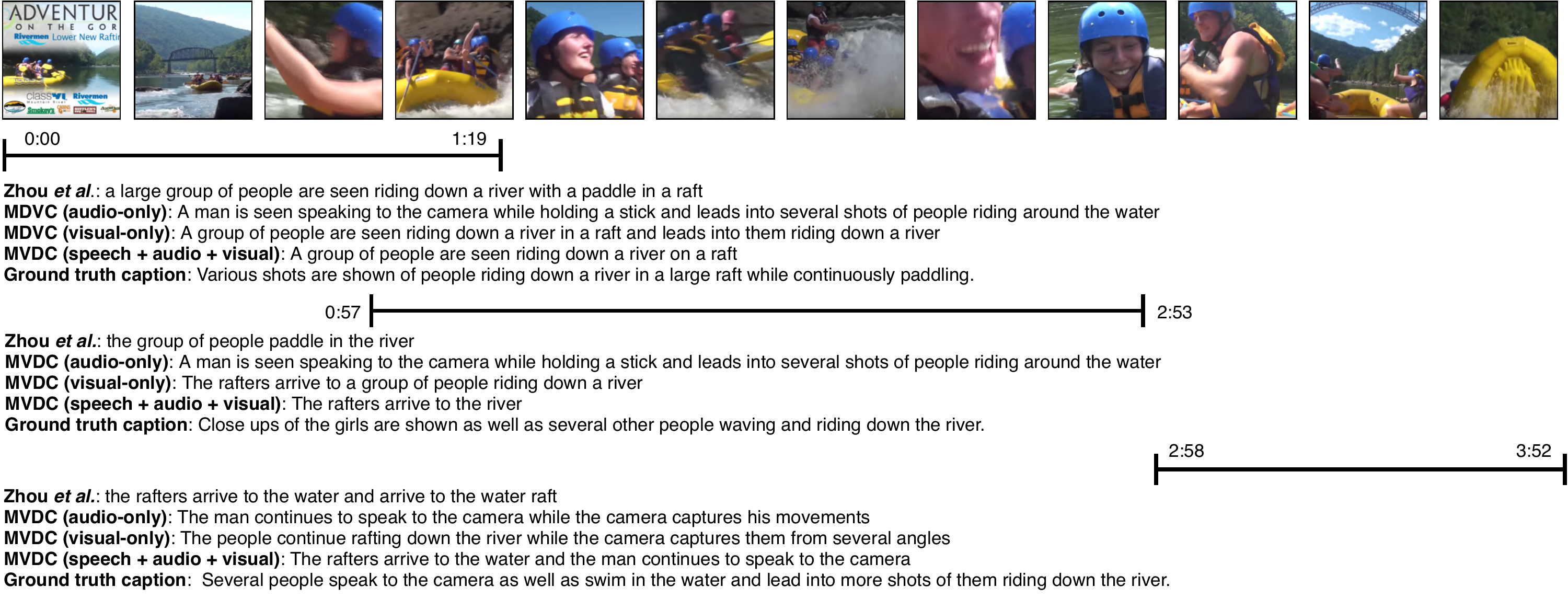}
    \end{center}
    \vspace{-1em}
    \caption{The qualitative captioning results for an example video from the ActivityNet Captions validation set. In the video, the speaker describes the advantages of rafting on this particular river and their club. Occasionally, people are shown rapturously speaking about how fun it is. Models that account for audio modality tend to grasp the details of the speaking on the scene while the visual-only models fail at this. We invite the reader to watch the example YouTube video for a better impression (\texttt{\href{https://www.youtube.com/embed/xs5imfBbWmw?rel=0}{xs5imfBbWmw}}).\label{fig:results_analysis} 
    }
\end{figure*}

\subsection{Ablation Studies}

\begin{figure}[t]
\begin{center}
\includegraphics[width=1\linewidth]{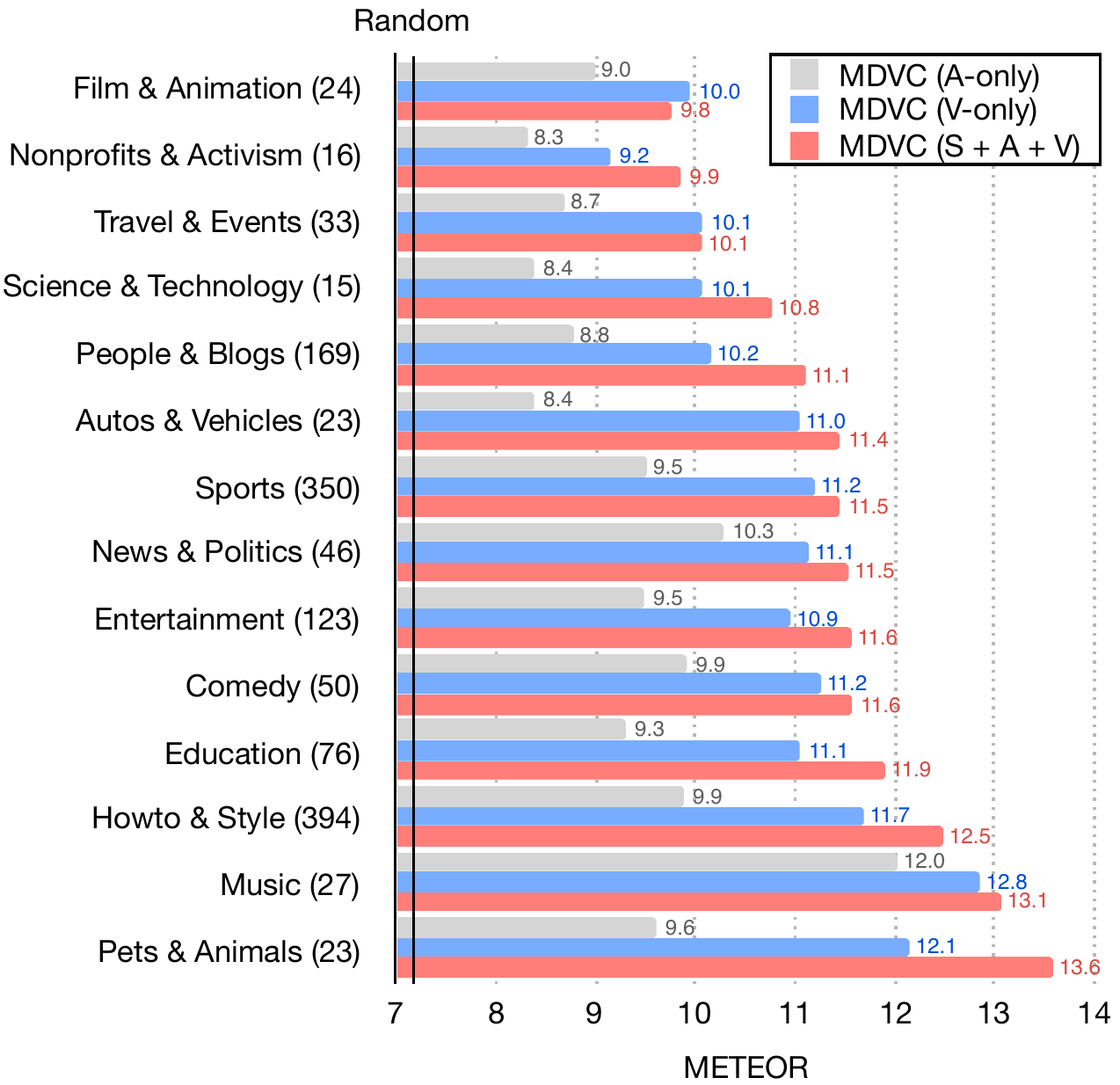}
\end{center}
\vspace{-1em}
   \caption{The results are split for category and version of MDVC. The number of samples per category is given in parenthesis. The METEOR axis is cut up to the random performance level (7.16).\label{fig:cat_chart2}}
\end{figure}

\begin{table}[t]
\centering
\begin{tabular}{lrrr}
\toprule
Model					  &Params.	& \multicolumn{2}{c}{Metric} \\
						& $(\times10^6)$ &B@4		&M \\  \midrule \midrule
Feature Transf. (random)&		42		&0.88	&7.16 \\
Bi-GRU					&		55		&1.44	&9.47 \\
Feature Transformer				&		42		&\textbf{1.84}	&\textbf{9.62} \\
\toprule
\end{tabular}
\vspace{0.5ex}
\caption{
 Comparison of the Feature Transformer and the Bi-directional GRU (Bi-GRU) architectures in terms of BLEU-4 (B@4), METEOR (M), and a number of model parameters. The input to all models is visual modality (I3D). The results indicate the superior performance of the Feature Transformer on all metrics. Additionally, we report the random input baseline which acts as a lower performance bound. The best results are highlighted.\label{table:ablation1}
}
\end{table}

\begin{table}[t]
\centering
\setlength\tabcolsep{0.5em}
\begin{tabular}{ccc r rrr}
\toprule
 \multicolumn{3}{c}{Modality} 	& \multirow{2}{*}{Fusion} &	Params. &	\multicolumn{2}{c}{Metric} \\
V & A	& S	& & $(\times10^6)$ & B@4 & M \\ \midrule \midrule
\ding{52}	& 	& 	& 	--	& 42	& 1.61	&9.64 \\
	& \ding{52}	& 	& 	--	& 5		& 1.03	&8.01 \\
\ding{52}	& \ding{52}	& 	& Average probs.	& 46	& 1.68	&9.71 \\
\ding{52}	& \ding{52}	& 	& Concat. + 2 FC	& 149	& 1.73	&9.87 \\
\ding{52}	& 	& 	& No, 2 FC	& 145	& 1.62	&9.69 \\
\ding{52}	& \ding{52}	& \ding{52}	& Concat. + 2 FC	& 179	& \textbf{1.81}	&\textbf{10.09} \\ \bottomrule
\end{tabular}
\vspace{1ex}
\caption{
The performance of the proposed MDVC framework with different input modalities (V-visual, A-audio, S-speech) and feature fusion approaches: probability averaging and concatenation of two fully-connected layers (Concat.~+ 2 FC). Also, we report the comparison between audio-visual MDVC with visual-only MDVC with similar model capacities (2 FC).\label{table:ablation2}
}
\end{table}

In this section, we perform an ablation analysis highlighting the effect of different design choices of our method. For all experiments, we use the full unfiltered ActivityNet Captions validation set with ground truth event proposals.

Firstly, we assess the selection of the model architecture. To this end, we implemented a version of our method where the transformer was replaced by \textit{Bidirectional Recurrent Neural Network with Gated Recurrent Units with attention} (Bi-GRU), proposed in \cite{Bahdanau2014}. To distil the effect of the change in architecture, the results are shown for visual-only models. Both Bi-GRU and the transformer input I3D features extracted from 64 RGB and optical flow frames (the final model inputs 24 frames). Finally, we set a lower bound for the feature performance by training a transformer model with random video features. Tab.~\ref{table:ablation1} shows the comparison. To conclude, we observe that the feature transformer-based model is not only uses less parameters but also achieves better performance in dense video captioning task. Moreover, both method clearly surpasses the random baseline. 

Secondly, we evaluate the contribution of different modalities in our framework. Tab.~\ref{table:ablation2} contains the results for different modality configurations as well as for two feature fusion approaches. Specifically, averaging of the output probabilities and concatenation of the outputs of all modalities and applying two fully connected (FC) layers on top. We observe that audio-only model has the worst performance, followed by the visual only model, and the combination of these two. Moreover, the concatenation and FC layers result in better performance than averaging. To further assess if the performance gain is due to the additional modalities or to the extra capacity in the FC layers, we trained a visual-only model with two additional FC layers. The results indicate that such configuration performs worse than any bi-modal setup. Overall, we conclude that the final model with all three modalities performs best among all tested set-ups, which highlights the importance of multi-modal setting in dense video captioning task.

Fig.~\ref{fig:results_analysis} shows a qualitative comparison between different models in our ablation study. Moreover, we provide the corresponding captions from the best performing baseline method (Zhuo \etal \cite{Zhou2018}). We noticed the following pattern: the audio-modality produces coherent sentences and captures the concepts of speaking in the video. However, there are clear mistakes in the caption content. In contrast, the model with all three modalities manages to capture the man who speaks to the camera which is also present in the ground truth. Both visual-only MDVC and Zhuo \etal struggle to describe the audio details.

Finally, to test whether our model improves the performance in general rather than in a specific video category, we report the comparison of the different versions of MDVC per category. To this end, we retrieve the category labels from the YouTubeAPI \cite{YouTubeCats} (US region) for every available ActivityNet Captions validation video. These labels are given by the user when uploading the video and roughly represent the video content type. The comparison is shown in Fig.~\ref{fig:cat_chart2}. The results imply a consistent gain in performance within each category except for categories: ``Film \& Animation'' and ``Travel \& Events'' which might be explained by the lack of correspondence between visual and audio tracks. Specifically, the video might be accompanied by music, \eg promotion of a resort. Also, ``Film \& Animation'' contains cartoon-like movies which might have a realistic soundtrack while the visual track is goofy.

\section{Conclusion}
The use of different modalities in computer vision is still an underrepresented topic and, we believe, deserves more attention. In this work, we introduced a multi-modal dense video captioning module (MDVC) and shown the importance of the audio and speech modalities for dense video captioning task. Specifically, MDVC is based on the transformer architecture which encodes the feature representation of each modality for a specific event proposal and produces a caption using the information from these modalities. The experimentation, conducted employing the ActivityNet Captions dataset, shows the superior performance of a captioning module to the visual-only models in the existing literature. Extensive ablation study verifies this conclusion. We believe that our results firmly indicate that future works in video captioning should utilize a multi-modal input.

{\footnotesize\noindent \textbf{Acknowledgments} Funding for this research was provided by the Academy of Finland projects 327910 \& 324346. The authors acknowledge CSC --- IT Center for Science, Finland, for computational resources.}

{\small
\bibliographystyle{ieee_fullname}
\bibliography{egbib}
}

\clearpage

\section{Supplementary Material}
The supplementary material consists of four sections. In Section~\ref{sec:another_example}, we provide qualitative results of the MDVC on another example video. The details on features extraction and implementation are described in Section~\ref{sec:feat_extraction} and \ref{sec:implementation_details}. Finally, the comparison with other methods is shown in Section~\ref{sec:other_methods}.

\subsection{Qualitative Results (Another Example)} \label{sec:another_example}
\begin{figure*}[t]
    \begin{center}
    \includegraphics[width=1\linewidth]{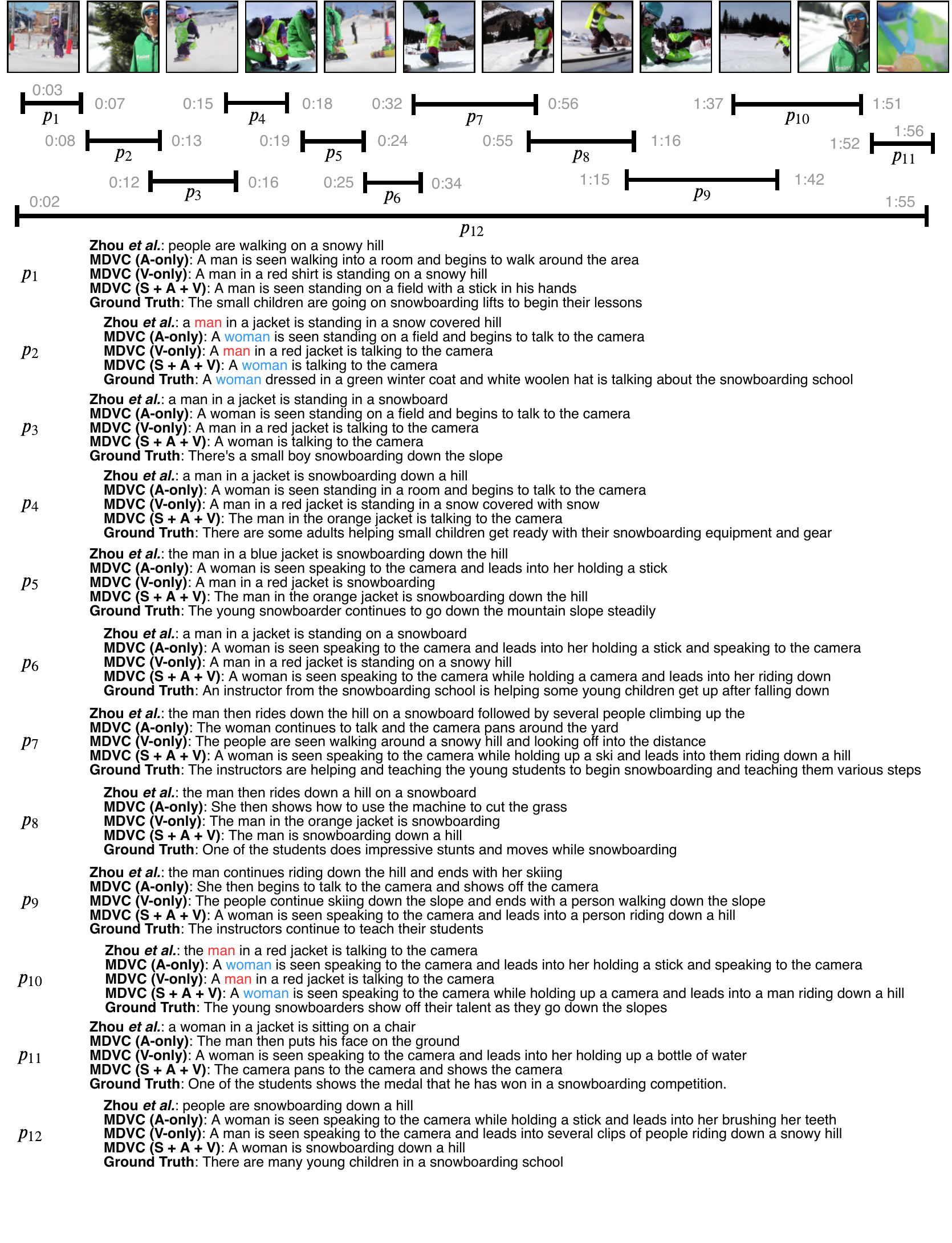}
    \end{center}
    \vspace*{-4.5em}
    \caption{Another example of the qualitative results for a video in the validation set. In the video, a lady is shown speaking twice (in $p_2$ and $p_{10}$). Since MDVC is conditioned not only on visual (V) but also speech (S) and audio (A) modalities, it managed to hallucinate a caption containing a ``\textcolor{blue}{woman}'' instead of a ``\textcolor{red}{man}''. We invite a reader to watch it on YouTube for a better impression  (\texttt{\href{https://www.youtube.com/embed/EGrXaq213Oc?rel=0}{EGrXaq213Oc}}). Note: the frame size mimics the MDVC input; the scale of temporal segments is not precise. Best viewed in color.}
    \label{fig:results_analysis2}
\end{figure*}

In Figure~\ref{fig:results_analysis2}, we provide qualitative analysis of captioning on another video from ActivityNet Captions validation set to emphasize the importance of additional modalities for dense video captioning, namely, speech and audio. We compare the captioning proposed by MDVC (our model) conditioned on different sets of modalities: audio-only (A-only), visual-only (V-only), and including all modalities (S + A + V). Additionally, we provide the results of a captioning model proposed in Zhou \etal \cite{Zhou2018} (visual only) which showed the most promising results according to METEOR.

More precisely, the video (YouTube video id: \texttt{\href{https://www.youtube.com/embed/EGrXaq213Oc?rel=0}{EGrXaq213Oc}}) lasts two minutes and contains 12 human annotations. The video is an advertisement for snowboarding lessons for children. It shows examples of children successfully riding a snowboard on a hill and supportive adults that help them to learn. A lady narrates the video and appears in the shot a couple of times.

Generally, we may observe that MDVC with the audio modality alone (A-only) mostly describes that a woman is speaking which is correct according to the audio content yet the details about snowboarding and children are missing. This is expectedly challenging for the network as no related sound effects to snowboarding are present. In the meantime, the visual-only MDVC grasps the content well, however, misses important details like the gender of the speaker. While the multi-modal model MDVC borrows the advantages of both which results in more accurate captions. The benefits of several modalities stand out in captions for $p_2$ and $p_{10}$ segments. Note that despite the appearance of the lady in the shot during $p_{10}$, the ground truth caption misses it yet our model manages to grasp it.

Yet, some limitations of the final model could be noticed as well. In particular, the content of some proposals is dissimilar to the generated captions, \eg the color of the jacket ($p_4$, $p_5$), or when a lady is holding a snowboard with a child on it while the model predicts that she is holding a ski ($p_7$). Also, the impressive tricks on a snowboard were guessed simply as ``ridding down a hill'' which is not completely erroneous but still inaccurate ($p_8$). Overall, the model makes reasonable mistakes except for proposals $p_3$ and $p_4$. Finally, the generated captions provide more general description of a scene compared to the ground truth that is detailed and specific which could be a subject for future investigation.

\subsection{Details on Feature Extraction} \label{sec:feat_extraction}
Before training, we pre-calculate the features for both audio and visual modalities. In particular, the audio features were extracted using VGGish \cite{Hershey2017} which was trained on AudioSet \cite{Gemmeke2017}. The input to the VGGish model is a $96\times64$ log mel-scaled spectrogram extracted for non-overlapping $0.96$ seconds segments. The log mel-scaled spectrogram is obtained by applying \textit{Short-Time Fourier Transform} on a $16$~kHz mono audio track using a periodic \textit{Hann} window with $25$ ms length with $10$ ms overlap. The output is a $128$-d feature vector after an activation function and extracted before a classification layer. Therefore, the input to MDVC is a matrix with dimension $T_j^a \times 128$ where $T_j^a$ is the number of features proposal $p_j$ consists of.

The visual features were extracted using I3D \cite{Carreira2017} network which inputs a set of $24$ RGB and optical flow frames extracted at $25$ fps. The optical flow is extracted with PWC-Net \cite{Sun2018PWC-Net}. First, each frame is  resized such that the shortest side is $256$ pixels. Then, the center region is cropped to obtain $224\times224$ frames. Both RGB and flow stacks are passed through the corresponding branch of I3D. The output of each branch are summed together producing $1024$-d features for each stack of $24$ frames. Hence, the resulting matrix has the shape: $T_j^v\times 1024$, where $T_j^v$ is the number of features required for a proposal $p_j$.

We use $24$ frames for I3D input to temporally match with the input of the audio modality as $\frac{24}{25} = 0.96$. Also note that I3D was pre-trained on the Kinetics dataset with inputs of $ 64 $ frames, while we use $ 24 $ frames. This is a valid approach since we employ the output of the second to the last layer after activation and average it on the temporal axis.

The input for speech modality is represented by temporally allocated text segments in the English language (one could think of them as subtitles). For a proposal $ p_j $, we pick all segments that both: a) end after the proposal starting point, \textit{and} b) start before the proposal ending point. This provides us with sufficient coverage of \textit{what has been said} during the proposal segment. Similarly to captions, each word in a speech segment is represented as a number which corresponds to the word's order number in the vocabulary and then passed through the text embedding of size $512$. We omit the subtitles that describe the sound like ``[Applause]'' and ``[Music]'' as we are only interested in the effect of the speech. Therefore, the speech transformer encoder inputs matrices of shape: $T^s_j\times 512$ where $T^s_j$ is the number of words in corresponding speech for proposal $p_j$.

\begin{table}[t!]
\centering
\begin{tabular}{lr}
\toprule
Method  & METEOR   \\ \midrule\midrule
\multicolumn{2}{l}{\textbf{\textit{Seen full dataset}}} \\
Xiong \cite{Xiong2018} (RL)          & 7.08  \\
Mun  \etal \cite{Mun2019} (RL) & 8.82  \\
Mun \etal \cite{Mun2019} (without RL) & 6.92  \\ \midrule \midrule
\multicolumn{2}{l}{\textbf{\textit{Seen part of the dataset}}} \\
MDVC, no missings & 7.31 \\ \bottomrule
\end{tabular}
\vspace{0.5ex}
\caption{
    The comparison with other dense video captioning methods on ActivityNet Captions validation set estimated with METEOR. The results are presented for the learned proposals.\label{tab:sota}
}
\end{table}

\subsection{Implementation Details} \label{sec:implementation_details}
Since no intermediate layers connecting the features and transformers are used, the dimension of the features transformers $D_T$ corresponds to the size of the extracted features: $512$, $128$, and $1024$ for speech, audio, and visual modalities, respectively. Each feature transformer has one layer ($L$), while the internal layer in the position-wise fully-connected network has $D_P=2048$ units for all modality transformers which was found to perform optimally. We use $H=4$ heads in all multi-headed attention blocks. The captions and speech vocabulary sizes are 10,172 and 23,043, respectively. 

In all experiments, except for the audio-only model, we use \textit{Adam} optimizer \cite{Kingma2014}, a batch containing features for $ 28 $ proposals, learning rate $10^{-5}$, $\beta = (0.9, 0.99)$, smoothing parameter $\gamma = 0.7$. In the audio-only model, we apply two-layered transformer architecture with learning rate $10^{-4}$ and $\gamma = 0.2$. To regularize the weights of the model, in every experiment, \textit{Dropout} \cite{Srivastava2014} with $p = 0.1$ is applied to the outputs of positional encoding, in every sub-layer before adding a residual, and after the first internal layer of the multi-modal generator. 

During the experimentation, models were trained for $200$ epochs at most and stopped the training early if for $50$ consecutive epochs the average METEOR score calculated on ground truth event proposals of both validation sets has not improved. At the end of the training, we employ the best model to estimate its performance on the learned temporal proposals. Usually the training for the best models culminated by $50^{\text{th}}$ epoch, \eg the final model (MDVC (S + A + V)) was trained for $ 30 $ epochs which took, roughly, $ 15 $ hours on \textit{one} consumer-type GPU (Nvidia GeForce RTX 2080 Ti). The code for training heavily relies on PyTorch framework and will be released upon publication.

\subsection{Comparison with Other Methods} \label{sec:other_methods}
In Tab.~\ref{tab:sota}, we present a comparison with another body of methods \cite{Xiong2018,Mun2019} which were not included in the main comparison as they were using \textit{Reinforcement Learning} (RL) approach to directly optimize the non-differentiable metric (METEOR). We believe that our method could also benefit from these as the ablation studies in \cite{Xiong2018,Mun2019} show significant improvement. As it was anticipated, in general, methods which employ reinforcement learning perform better in terms of METEOR. Interestingly, our model still outperforms \cite{Xiong2018} which uses RL in the captioning module.

\end{document}